\documentclass[
]{ceurart}

\usepackage{lipsum}
\usepackage{todonotes}

\begin{document}

\copyrightyear{2023}
\copyrightclause{Copyright for this paper by its authors.
  Use permitted under Creative Commons License Attribution 4.0
  International (CC BY 4.0).}

\conference{ETAPS2024, Europeans joint conferences on theory and practice of software, International Workshop on Reliability Engineering Methods for Autonomous Robots - REMARO 2024, Apr 6, Luxembourg City, Luxembourg}

\title{Knowledge Distillation in YOLOX-ViT for Side-Scan Sonar Object Detection}
\author[1]{Martin Aubard}[%
orcid=0009-0000-3070-8067,
email=maubard@oceanscan-mst.com,
]
\address[1]{OceanScan Marine Systems \& Technology, 4450-718 Matosinhos, Portugal}

\author[2]{László Antal}[%
orcid=0009-0005-4977-0959,
email=antal@informatik.rwth-aachen.de,
]
\address[2]{RWTH Aachen University, 52074 Aachen, Germany}

\author[3]{Ana Madureira}[%
orcid=0000-0002-0264-4710,
email=amd@isep.ipp.pt,
]
\address[3]{Interdisciplinary Studies Research Center (ISRC), ISEP/IPP, 4200-485 Porto, Portugal}

\author[2]{Erika Ábrahám}[%
orcid=0000-0002-5647-6134,
email=abraham@informatik.rwth-aachen.de ,
]


\begin{abstract}
In this paper we present YOLOX-ViT, a novel object detection model, and investigate the efficacy of knowledge distillation for model size reduction without sacrificing performance. Focused on underwater robotics, our research addresses key questions about the viability of smaller models and the impact of the visual transformer layer in YOLOX. Furthermore, we introduce a new side-scan sonar image dataset, and use it to evaluate our object detector's performance. Results show that knowledge distillation effectively reduces false positives in wall detection. Additionally, the introduced visual transformer layer significantly improves object detection accuracy in the underwater environment. The source code of the knowledge distillation in the YOLOX-ViT is at \url{https://github.com/remaro-network/KD-YOLOX-ViT}.
\end{abstract}



\maketitle

\section{Introduction}
\label{sec:introduction}

\begin{wrapfigure}[12]{r}{8cm}
    \centering
    \vspace*{-0.50cm}
    \includegraphics[width=\linewidth]{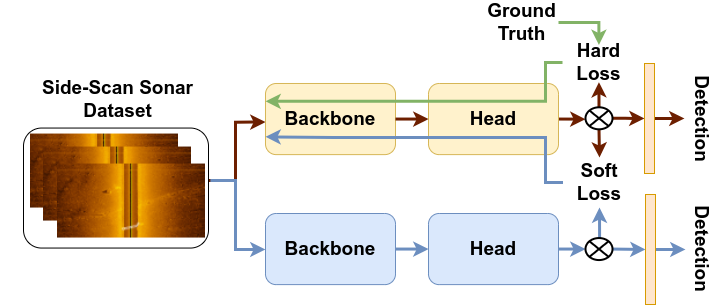}
    \vspace*{-0.75cm}
    \caption{Knowledge distillation: blue nodes indicate the teacher, while yellow ones show the student. The combined hard and soft loss is used to train the student network.}
    \label{fig:yolox-kd}
\end{wrapfigure}

Over the past few decades, there has been a growing interest in exploring oceanic environments, leading to an expansion of underwater activities such as infrastructure development \cite{WU2019379} and archaeological explorations \cite{https://doi.org/10.1002/rob.20350}.
Due to the unpredictable and often unknown nature of underwater conditions, autonomous underwater vehicles (AUVs) have become pivotal in performing tasks ranging from surveying to maintenance.
AUVs enable data collection and execution of underwater operations based on predefined plans.
Nevertheless, the complexity of the underwater environment requires enhanced decision making capabilities, and thus high understanding of the vehicle's environment. 
Computer vision based on deep learning (DL) offers a promising solution for embedded real-time detection in such scenarios \cite{AubardSSS}. Yet, the substantial size of standard DL models presents significant challenges for AUVs regarding power consumption, memory allocation, and the need for real-time processing.
Recent research has concentrated on model reduction to facilitate the onboard implementation of DL models on embedded systems \cite{Chen2017distilling}.
This paper contributes to this body of work by:
\begin{itemize}
    \item Improving the state-of-the-art object detection model, YOLOX, by integrating a vision transformer layer, resulting in the YOLOX-ViT model.
    \item Implementing a knowledge distillation (KD) method to improve the performance of smaller YOLOX-ViT models, such as YOLOX-Nano-ViT (see Figure \ref{fig:yolox-kd}).
    \item Introducing a novel side-scan sonar (SSS) dataset specifically for wall detection.
\end{itemize}

This paper is organized as follows: Section \ref{sec:related-work} reviews related work on object detection and knowledge distillation. Section \ref{sec:model} presents the YOLOX and YOLOX-ViT models and introduces a new side-scan sonar dataset. Section \ref{sec:knowledge-distillation} explains how we applied knowledge distillation. Section \ref{sec:exp-results} discusses the experiments and their results. Finally, Section \ref{sec:conclusion} concludes the paper.

\section{Related Work}
\label{sec:related-work}




\textbf{Object detection.} Object detection is a basic task in computer vision (CV) with the goal of detecting visual objects of certain classes in digital images \cite{10028728}.
Its aim is to develop efficient algorithms providing the essential information for CV applications: \textit{"What objects are where?"}.
Their quality is measured by detection accuracy (classification and localization) and efficiency (prediction time).
These methods seek a practical trade-off between accuracy and efficiency.
The increasing popularity of deep learning \cite{lecun2015deep, zhao2019object} has led in remarkable breakthroughs in object detection.
Thus, it is widely used in many real-world applications, such as autonomous driving \cite{chen2017multi}, robot vision \cite{wan2020faster}, and video surveillance \cite{nascimento2006performance}.
Object detectors based on deep learning fall into two categories: "two-stage detectors" and "one-stage detectors" \cite{10028728}. Notably, You Only Look Once (YOLO) is a representative one-stage detector \cite{Redmon_2016_CVPR}, known for its speed and effectiveness \cite{zhiqiang2017review, JIANG20221066}. YOLO's simplicity lies in its ability to directly output bounding box position and category through the neural network \cite{JIANG20221066}. Over the past years, various YOLO versions such as YOLO V2 \cite{DBLP:journals/corr/RedmonF16}, YOLO V3 \cite{DBLP:journals/corr/abs-1804-02767}, YOLO V4 \cite{DBLP:journals/corr/abs-2004-10934}, YOLO V5 \cite{YADAV2022292} and YOLOX \cite{DBLP:journals/corr/abs-2107-08430}, have been introduced.
YOLOX, an anchor-free version (detailed explanation provided in \autoref{sec:yolox}), follows a simpler design with superior performance, aiming to bridge the gap between research and industry \cite{DBLP:journals/corr/abs-2107-08430}. 

\noindent \textbf{Knowledge Distillation.} As explained in Section \ref{sec:introduction}, onboard DL models must satisfy real-life constraints such as efficiency and accuracy. 
Larger models typically yield better results due to their ability to learn complex relations, but they come with higher computational resource consumption. 
Research on reducing model size while maintaining performance has been ongoing, including model pruning focused on removing less critical neurons or weights \cite{molchanov2017pruning, NIPS2015_ae0eb3ee}.

G. Hinton et al. \cite{hinton2015distilling} introduced knowledge distillation (KD) for image classification, transferring knowledge from a well-trained larger model (teacher) to a smaller one (student) using a new loss function.
This loss function combines the \textit{ground truth loss} $\mathcal{L}_{\text{hard}}$ and \textit{distillation loss} $\mathcal{L}_{\text{soft}}$, where the former guides predictions towards ground truth labels, and the latter leverages knowledge from a larger model to align the behavior of the smaller one. 
The loss function is expressed as $\mathcal{L} = \lambda \cdot \mathcal{L}_{\text{hard}} + (1 - \lambda) \cdot \mathcal{L}_{\text{soft}}$, with $\lambda$ as a parameter regularizing the hard and soft loss terms.

Chen et al. \cite{Chen2017distilling} demonstrated KD's effectiveness in compressing single-shot detector (SSD) models, introducing an improved method using a "hint" loss function, based on intermediate backbone logits. 
Wei et al. \cite{wei2018quantization} proposed a novel approach where the student learns from both the teacher model's final output and intermediate feature representations, improving performance in complex scenes.
Zhang et al. \cite{Zhang_2015} tailored KD techniques for Faster R-CNN, achieving efficiency gains without compromising detection accuracy. Q. Yang et al. \cite{Yang2022} applied KD to YOLOv5, using YOLOv5-m as teacher and the YOLOv5-n as student for bell detection, which improved the YOLOv5-n $\text{mAP}_{50}$ about $2\%$.


\section{YOLOX-ViT}
\label{sec:model}


\begin{figure}
    \centering
    \includegraphics[width=1\linewidth]{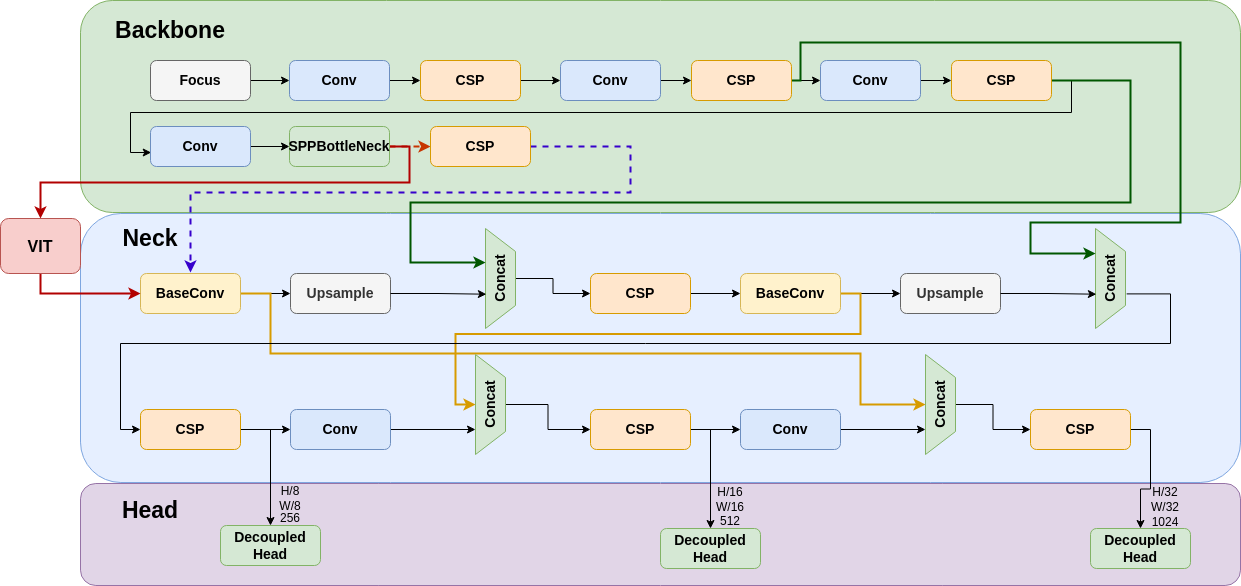}
    \caption{The YOLOX-ViT model with its three main parts: i. backbone for the feature extraction, ii. neck characterized by a feature pyramid network (FPN) connecting the backbone and the head, and iii. (decoupled) head for the bounding box regression and classification tasks. The Visual Transformer layer (ViT) is located between the backbone and the neck represented by the red arrow, in contrast the basic YOLOX architecture is represented by the blue dotted line. For further explanation of the individual blocks, we refer to Figure \ref{fig:yolo-blocks} of the appendix.}
    \label{fig:yolox-vit}
\end{figure}

Transformers, introduced by Vaswani et al. \cite{vaswani2023attention}, initially designed for natural language processing, proved effective in handling sequential data, outperforming the state-of-the-art.
Dosovitskiy et al. \cite{dosovitskiy2021image} introduced the Visual Transformer called ViT, the first computer vision transformer model, achieving state-of-the-art performance on image recognition tasks without convolutional layers. Carion et al. \cite{carion2020endtoend} presented DETR (DEtection TRansformer) for object detection directly predicting sets (bounding boxes, class labels, and confidence score), without the need of separate region proposal and refinement stages. Further description of the ViT can be found in Section \ref{sec:transformers} of the Appendix. Integrating transformers with CNNs enhances feature extraction in object detection tasks, combining the spatial hierarchy of CNNs with the global context of transformers. Yu et al. \cite{rs13183555} proposed a YOLOv5-TR for containers and shipwreck detection. Aubard et al. \cite{AubardSSS} demonstrated a 5.5\% performance improvement using YOLOX. Our paper suggests enhancing the YOLOX model by incorporating a transformer layer (see Figure \ref{fig:yolox-vit}) between the backbone and neck at the end of the \textit{SPPBottleneck} layer for improved feature extraction capability. The ViT layer is set up with 4 multi-head self-attention (MHSA) sub-layers.

\subsection{Underwater SSS Image Dataset}
\label{subsec:datatset}

The Sonar Wall Detection Dataset (SWDD) consists of side-scan sonar (SSS) imagery collected in Porto de Leixões harbor using a Klein 3500 sonar on a light autonomous underwater vehicle (LAUV) \cite{SOUSA2012268} operated by OceanScan Marine Systems \& Technology. The survey produced 216 images along harbor walls. Due to the small amount of data in the current dataset, a data augmentation method is used to increase the data and improve the robustness of the model. The data augmentation used in this paper introduced noise, flips, and combined noise-flip transformations. 
\\
The noise is a Gaussian noise represented as $I_{\text{noisy}}(x, y) = I(x, y) + N(0, \sigma^2)$. Where $I_{\text{noisy}}(x, y)$ is the intensity value of the pixel at position $(x, y)$ in the noisy image, $I(x, y)$ is the original intensity at position $(x, y)$, and $N(0, \sigma^2)$ represents a sample from a Gaussian distribution with mean 0 and variance $\sigma^2$. The standard deviation $\sigma$ for this dataset equals $0.5 \times 255$.
\\
The horizontal flip described as $I_{\text{flipped}}(x, y) = I(W - x - 1, y)$. It mirrors the image across its vertical axis, where $W$ is the image's width, $(W - x - 1)$ effectively changes the horizontal position of each pixel.
\\
The dataset features "Wall" and "NoWall" classes, with $2.616$ labeled samples. We combined original and augmented images for robust training, partitioning the dataset into $70\%$ for training, $15\%$ for validation, and $15\%$ for testing. Each SSS image, generated from a 75m range on each side with a frequency of 900 kHz, consists of 500 consecutive lines, resulting in a resolution of $4.168 \times 500$ pixels. We scaled the images to $640 \times 640$ for compatibility with computer vision algorithms. The SWDD is publicly accessible for research purposes under \url{https://zenodo.org/records/10528135}.
\section{Knowledge Distillation}
\label{sec:knowledge-distillation}

Considering YOLOX's architecture, KD is applied to each decoupled head of the feature pyramid network (FPN), as outlined in Section \ref{sec:related-work}. Due to this architectural particularity, the KD process involves computing distinct loss functions for each FPN output, encompassing classification, bounding box regression, and objectness scores between the student and the teacher model. 
The general KD loss function is defined as:
$\mathcal{L}_{\text{KD}} = \mathcal{L}_{\text{hard}} + \mathcal{L}_{\text{soft}}$,
where
$\mathcal{L}_{\text{soft}} = \lambda_{\text{bbox}} \cdot \mathcal{L}_{\text{KD\_bbox}} + \lambda_{\text{obj}} \cdot \mathcal{L}_{\text{KD\_obj}} + \lambda_{\text{cls}} \cdot \mathcal{L}_{\text{KD\_cls}}$
(detailed explanation provided in \autoref{sec:KD_Losses}).
For each type of loss (classification, objectness, bounding box), the loss function is decomposed across the outputs of each FPN, resulting in the following formulations:

\textbf{Classification Loss:}
\begin{equation}
\mathcal{L}_{\text{KD\_cls}} = \frac{\mathcal{L}_{\text{KD\_cls\_FPN0}} + \mathcal{L}_{\text{KD\_cls\_FPN1}} + \mathcal{L}_{\text{KD\_cls\_FPN2}}}{batch\_size \times 3}
\end{equation}

\textbf{Bounding Box Regression Loss:}
\begin{equation}
\mathcal{L}_{\text{KD\_bbox}} = \frac{\mathcal{L}_{\text{KD\_bbox\_FPN0}} + \mathcal{L}_{\text{KD\_bbox\_FPN1}} + \mathcal{L}_{\text{KD\_bbox\_FPN2}}}{batch\_size \times 3}
\end{equation}

\textbf{Objectness Loss:}
\begin{equation}
\mathcal{L}_{\text{KD\_obj}} = \frac{\mathcal{L}_{\text{KD\_obj\_FPN0}} + \mathcal{L}_{\text{KD\_obj\_FPN1}} + \mathcal{L}_{\text{KD\_obj\_FPN2}}}{batch\_size \times 3}
\end{equation}


In our experiments, the $\mathcal{L}_{\text{KD}}$ is formulated to incorporate both hard loss and soft loss components with the soft loss terms ($\mathcal{L}_{\text{KD\_cls}}$, $\mathcal{L}_{\text{KD\_bbox}}$, and $\mathcal{L}_{\text{KD\_obj}}$) are each scaled by coefficients of $0.5$, i.e., $\lambda_{\text{bbox}} = \lambda_{\text{obj}} = \lambda_{\text{cls}} = 0.5$. 
This approach ensures that while the soft loss components are crucial for transferring nuanced knowledge from the teacher model to the student model, the hard loss remains predominant in guiding the student model's learning process, ensuring that it does not drift too far from the fundamental task objectives.

Moreover, this paper uses the YOLOX-L model as teacher and the YOLOX-Nano as student.
Because of the YOLOX's online random data augmentation, the online KD procedure is: \textbf{i.} train YOLOX-L, \textbf{ii.} initiate YOLOX-Nano training, \textbf{iii.} for each batch, run inference with YOLOX-L, \textbf{iv.} compute $\mathcal{L}_{\text{soft}}$ between FPN logits of YOLOX-L and YOLOX-Nano, and \textbf{v.} combine $\mathcal{L}_{\text{soft}}$ with $\mathcal{L}_{\text{hard}}$. This online KD approach takes a week due to the need for inference at each iteration.
This paper adopts an offline KD method without random online data augmentation during YOLOX-Nano training to reduce the time effort. The workflow is: \textbf{i.} train YOLOX-L, \textbf{ii.} perform YOLOX-L inference, saving FPN logits for each training sample, \textbf{iii.} launch YOLOX-Nano training, \textbf{iv.} retrieve corresponding FPN logits for each image, \textbf{v.} calculate $\mathcal{L}_{\text{soft}}$ between logits of each FPN output from YOLOX-L and YOLOX-Nano, and \textbf{vi.} combine $\mathcal{L}_{\text{soft}}$ with $\mathcal{L}_{\text{hard}}$.


\section{Experimental Evaluation}
\label{sec:exp-results}


\begin{table}
  \caption{Validation Results: \textit{TP} and \textit{FP} are, respectively, the True and False positive percentages over the extracted video images. \textit{Pr} is the precision of the detection based on the proportion of correctly identified objects among all detections, with precise bounding box placement. In contrast, \textit{Average Precision at 50\% IoU ($\textit{AP}_{50}$)} evaluates the balance between \textit{precision} and \textit{recall} (proportion of actual objects correctly identified by the model, again considering the accuracy of the bounding box positioning), thereby providing a comprehensive assessment of the model's ability to accurately detect objects, ensure correct bounding box alignment, and minimize false positives and negatives. $\textit{AP}_{50}$ average precision at an Intersection over Union (IoU) threshold of 0.5 where the \textit{AP} considers a range of IoU thresholds from 0.5 to 0.95. \textit{Detection} on video quantifies the duration, manually timed, where the model correctly detects a wall when it is in the video. \textit{FP} on video denotes the count of false positive detections, indicating instances where the model incorrectly identified a wall during the video inference. Further explanation of the metrics calculation can be found in Appendix \ref{sec:eval_metrics}.}
  \begin{tabular}{lccccc|cc}
    \toprule
    \textit{Model}         & \textit{TP} & \textit{FP} & \textit{Pr} & $\textit{AP}_{50}$        & \textit{AP}      & \textit{Detection}  & \textit{FP}        \\
    \midrule
    L       & 12.27 & 55          & 18.24         & 0.18          & 0.043          & 81.58          & 1.22          \\
    L-ViT    & 15.58 & 58.7          & 20.97         & 0.20 & 0.060 & 89.93          & 1.92         \\
    Nano    & 12.3 & 26.7           & 31.53         & 0.19          & 0.064          & 66.21          & 3.81          \\
    Nano-ViT & 24.9 & 36.9   & 40.29   & 0.42          & 0.15          & 83.93          & 0.25 \\
    \bottomrule
    L-noAug     & 26.33 & 49.66          & 34.64         & 0.33 & 0.13& 87.05          & 9.35          \\
    L-ViT-noAug    & 30.15 & 50.93       & 37.18         & 0.41     & 0.16          & 98.93          & 7.76          \\
    Nano-noAug    & 28.34          & 65.83 & 30.09         & 0.37            & 0.13             & 93.53             & 24.70          \\
    Nano-ViT-noAug    & 29.4 & 78.48  & 27.25          & 0.38        & 0.13          & 97.41          & 34.25          \\
    \bottomrule
  \end{tabular}
  \label{tab:val}
\end{table}

Besides the initial results obtained during training, we implemented an online validation process to assess model performance on real-world data. This evaluation centered around a video lasting 6 minutes and 57 seconds from a distinct survey. In pursuit of a thorough assessment, we meticulously timed and manually recorded both true and false positive detections during video analysis. Furthermore, we extracted and annotated frames from the video for a more precise evaluation of model accuracy, resulting in a total of 6,243 annotated images. Subsequently, the prediction bounding boxes were saved for each model's inference on the video. Utilizing the \textit{Object Detection Metrics} \cite{githubrepoVal} repository, we calculated metrics such as \textsf{\textit{TP}}, \textsf{\textit{FP}}, \textsf{\textit{Pr}} (precision), $\textsf{\textit{AP}}_{50}$, and \textsf{\textit{AP}}.
This evaluation process was applied to both YOLOX-L and YOLOX-Nano models in their standard architectures and the YOLOX-ViT variants. The results of this validation are presented in Table \ref{tab:val}. The experiment is divided into two parts, as detailed in Section \ref{sec:knowledge-distillation}. The first part involves training with online random data augmentation, while the second part proceeds without it.
The comparative analysis between the YOLOX-L and YOLOX-L-ViT models, particularly in the experiments with extracted images, reveals some limitations. YOLOX-L and YOLOX-L-ViT models underperformed compared to their counterparts without online data augmentation (YOLOX-L-noAug and YOLOX-L-ViT-noAug). This discrepancy suggests that removing online data augmentation yields better model performance, and the constraints of a limited dataset size might disproportionately affect the efficiency of larger models such as YOLOX-L. This assumption is further supported by an observed tendency towards overfitting in the augmented models and validated by visual experiments. These experiments indicated that while YOLOX-L-noAug manages to sustain detection for longer durations than YOLOX-L, it also significantly increases false positives by $8.13\%$. Furthermore, visual experiments demonstrate that YOLOX-L-ViT offers superior detection performance.
Additionally, as shown in Table \ref{tab:val}, integrating the ViT layer enhances the basic YOLOX model's capabilities. It boosts video detection by approximately $8\%$ with L, and $23\%$ for Nano. However, visual interpretations indicate that the ViT-enhanced models underperform relative to their basic architecture without online data augmentation. While detection rates increased, so did false positives, suggesting that the models may overly generalize bright objects as walls. 

The objective is to utilize knowledge distillation from L and L-ViT as 'teachers' to reduce the \textit{False Positive} rate in the smaller models Nano-noAug and Nano-ViT-noAug, designated as 'students'. Table \ref{tab:KD_val} demonstrates that KD effectively decreased the \textit{False Positive} rate. Additionally, the ViT layer further reduced \textit{False Positives} in the student models. Specifically, for Nano-noAug, the basic model lowered \textit{False Positives} by about 0.3\%, whereas the ViT variant reduced by $\sim$6\%. The Nano-ViT-noAug, with the basic model as a teacher, cut \textit{False Positives} by $\sim$12.95\%, and the ViT version managed a reduction of about 20.35\%. These results underscore the ViT layer's ability to aid the model in focusing on more effective feature extraction.

\begin{table*}
  \caption{Knowledge Distillation Results.}
  \label{tab:KD_val}
  \begin{tabular}{llccccc|cc}
    \toprule
    \textit{Teacher}         & \textit{Student} & \textit{TP} & \textit{FP} & \textit{Pr} & $\textit{AP}_{50}$        & \textit{AP}      & \textit{Detection}  & \textit{FP}        \\
    \midrule
    L-ViT    & Nano-noAug          & 32.4         & 75.9     & 29.91      & 0.35 & 0.14  & 96.6          & 18.7 \textbf{(-6)}       \\
    L    & Nano-noAug          & 29.13 & 84.92          & 25.54          & 0.30  & 0.11          & 97.6          & 24.4 (-0.3)         \\
    \bottomrule
    L-ViT    & Nano-ViT-noAug          & 28.9     & 66.92      & 30.16     & 0.35 & 0.13 & 95.7          & 13.9 \textbf{(-20.35)}          \\
    L    & Nano-ViT-noAug          & 29.79 & 60.10     & 33.31      & 0.35 & 0.14      & 94.7          & 21.3  (-12.95)          \\
 
    \bottomrule
  \end{tabular}
\end{table*}

 




\section{Conclusion}
\label{sec:conclusion}

In conclusion, our research presents a framework for advancing object detection in side-scan sonar images within underwater robotics and autonomous underwater vehicles. The introduction of YOLOX-ViT, coupled with applying knowledge distillation techniques, has proven effective in reducing model size while preserving high detection accuracy.

Our methodology yielded promising results by integrating visual transformer layers into the YOLOX backbone and employing knowledge distillation across decoupled heads. Specifically, knowledge distillation successfully alleviated false positives in wall detection, showcasing its practical utility in enhancing the efficiency of object detection models. Additionally, the integration of the ViT layer demonstrated a substantial improvement in overall accuracy, underscoring the potential of this architectural enhancement for underwater image analysis.

As autonomous underwater vehicles play a pivotal role in exploration and surveillance, ensuring the reliability and robustness of object detection models becomes essential. Our future efforts will focus on \textbf{i.} increasing the SSS dataset to improve the L metrics, \textbf{ii.} implementing KD with online data augmentation, and \textbf{iii.} developing methodologies for comprehensive safety verification, addressing potential challenges and uncertainties in real-world underwater scenarios. By refining the model based on safety-driven criteria, we aim to elevate the practical utility of our object detection framework, contributing to the broader objective of deploying AUVs in environments where precision and safety are equally critical considerations.

\bigskip

\noindent \textbf{Acknowledgments.} This project has received funding from the European Union's Horizon 2020 research and innovation programme under the Marie Skłodowska-Curie grant agreement No. 956200. For more info, please visit https://remaro.eu.

\bibliography{sample-ceur}

\newpage

\appendix
\section{Evaluation Metric Definitions}
\label{sec:eval_metrics}
List of abbreviations:
\begin{itemize}
    \item \textsf{\textit{TP}}: number of true positives
    \item \textsf{\textit{FP}}: number of false positives
    \item \textsf{\textit{PR}}: precision
    \item $\textsf{\textit{AP}}_{\text{50}}$: average precision at 50\% IoU
    \item \textsf{\textit{AP}}: average precision from 0.5\% to 0.95\% IoU 
    \item \textsf{\textit{MSE}}: mean squared error
    \item \textsf{\textit{BCE}}: binary cross entropy
    \item \textsf{\textit{KL\_div}}: Kullback-Leibler divergence
\end{itemize}

\subsection{Function definitions}

\noindent \textbf{Softmax.}
\begin{equation}
    \text{softmax}(z) = \frac{e^{z_i}}{\sum_{j=1}^{n}e^{z_j}} 
\end{equation}
Where \( e^{z_i} \) is the exponential of the \( i \)-th element of the input vector \( z \), and the denominator is the sum of exponentials of all elements in \( z \). This ensures that the softmax output is a probability distribution, with each element ranging between 0 and 1 . 
\\

\noindent \textbf{LogSoftmax.}
\begin{equation}
    \text{log\_softmax}(z) = \log{(\frac{e^{z_i}}{\sum_{j=1}^{n}e^{z_j}})}
\end{equation}
Natural logarithm of the softmax function.
\\

\noindent \textbf{Mean Squared Error (MSE.)}
It measures the error quantification of a model in regression tasks, by calculating the average of the squares of the differences between the predicted and actual values. The MSE equation is as follows:

\begin{equation}
    \text{MSE} = \frac{1}{N} \sum_{i=1}^{N} (y_i - \hat{y}_i)^2
\end{equation}
where \(y_i\) is the actual value, \(\hat{y}_i\) represent the predicted value, and \(N\) is the number of observations. 
\\

\noindent \textbf{Binary Cross-Entropy (BCE).}
It measures the classification model performance between 0 and 1. BCE equation is as follows:

\begin{equation}
    \text{BCE} = -\frac{1}{N} \sum_{i=1}^{N} [y_i \log(p_i) + (1 - y_i) \log(1 - p_i)]
\end{equation}
where \(y_i\) is the actual label (0 or 1), \(p_i\) is the predicted probability of the instance being in class 1, and \(N\) is the number of samples.
\\

\noindent \textbf{Kullback-Leibler Divergence (KL Divergence).}
The Kullback-Leibler Divergence measures how one probability distribution diverges from a second reference probability distribution. KL Divergence equation is as follows:

\begin{equation}
    \text{KL Divergence} = \sum_{i} P(i) \log\left(\frac{P(i)}{Q(i)}\right)
\end{equation}
\(P\) and \(Q\) are the two probability distributions. The KL Divergence measures the information lost when \(Q\) is used to approximate \(P\); a lower value indicates a closer match between the distributions.

\subsection{Model evaluation metrics}



\noindent \textbf{Precision.}
\begin{equation}
    \textit{\textsf{Pr}} = \frac{\textit{\textsf{TP}}}{\textit{\textsf{TP}} + \textit{\textsf{FP}}}
\end{equation}
Precision is a key performance metric in object detection and classification tasks. It quantifies the accuracy of positive predictions. \textit{\textsf{TP}} and \textit{\textsf{FP}} are respectively True Positive and False Positive detection. A higher Precision value indicates a lower rate of false positives, signifying that it is more likely to be correct when the model predicts a positive class.
\\










\noindent \textbf{AP50.}
\begin{equation}
    \textit{\textsf{AP}}_{50} = \int_{0}^{1} p_{50}(r) \, dr
\end{equation}
$\textit{\textsf{AP}}_{50}$ represents the Average Precision at an IoU threshold of 50\%. It is calculated as the area under the precision-recall curve for an IoU 50\%. Here, \(p_{50}(r)\) denotes the precision at a recall level \(r\) for IoU = 50\%, effectively integrating precision overall recall levels.
\\

\noindent \textbf{AP.}
\begin{equation}
    \textit{\textsf{AP}} = \frac{1}{T} \sum_{t \in \{50, 55, \hdots, 95\}} \textit{\textsf{AP}}_t
\end{equation}
\textit{\textsf{AP}} (Average Precision) is an aggregate measure over multiple IoU thresholds, comprehensively evaluating object detection performance. It is the mean of \textit{\textsf{AP}} computed at each IoU threshold \(t\) from 50\% to 95\%, where \(T\) is the total number of thresholds considered.
\\

\noindent \textbf{Intersection over Union (IoU).}
Metric measuring the accuracy of an object detector on a particular dataset. It is particularly known to evaluate object detection and segmentation models. IoU equation is as follows:

\begin{equation}
    \text{IoU} = \frac{\text{Area of Overlap}}{\text{Area of Union}}
\end{equation}

The formula represents the intersection area between the predicted and ground truth bounding box. IoU ranges from 0 to 1, where 1 indicates a perfect match between the predicted bounding box and the ground truth. A higher IoU signifies a higher accuracy of the object detection model.

\subsection{Knowledge Distillation Losses}
\label{sec:KD_Losses}

\noindent \textbf{KD Bounding Box Loss.}
\begin{equation}
    \mathcal{L}_{\text{KD\_bbox}} = \frac{1}{N} \sum_{i=1}^{N} \textit{\textsf{{MSE}}}(\text{S\_reg}_i, \text{T\_reg}_i)
\end{equation}
This formula calculates the Knowledge Distillation loss for bounding box regression. It employs the Mean Squared Error (MSE) to compare the bounding box regression outputs (\(\text{S\_reg}_i\)) from the student model with those from the teacher model (\(\text{T\_reg}_i\)), averaged over \(N\) data samples.
\\

\noindent \textbf{KD Object Loss.}
\begin{equation}
    \mathcal{L}_{\text{KD\_obj}} = \frac{1}{N} \sum_{i=1}^{N} \textit{\textsf{{BCE}}}(\text{S\_obj}_i, \text{T\_obj}_i)
\end{equation}
This equation computes the KD loss for objectness predictions using Binary Cross-Entropy (BCE). It evaluates the difference between the objectness scores predicted by the student (\(\text{S\_obj}_i\)) and the teacher (\(\text{T\_obj}_i\)) models.
\\

\noindent \textbf{KD Class Loss.}
\begin{equation}
    \mathcal{L}_{\text{KD\_cls}} = \frac{1}{N} \sum_{i=1}^{N} \textit{\textsf{{KL\_div}}}(\text{log\_softmax}(\text{S\_cls}_i), \text{softmax}(\text{T\_cls}_i))
\end{equation}
Here, the loss for class predictions in KD is quantified. The Kullback-Leibler (KL) divergence measures the disparity between the log-softmax outputs of the student's class predictions (\(\text{S\_cls}_i\)) and the softmax outputs from the teacher (\(\text{T\_cls}_i\)).
\\

\noindent \textbf{Overall Soft Loss.}
\begin{equation}
    \mathcal{L}_{\text{soft}} = \lambda_{\text{bbox}} \cdot \mathcal{L}_{\text{KD\_bbox}} + \lambda_{\text{obj}} \cdot \mathcal{L}_{\text{KD\_obj}} + \lambda_{\text{cls}} \cdot \mathcal{L}_{\text{KD\_cls\_loss}}
\end{equation}
The overall soft loss for the KD process is a weighted sum of the individual KD losses (bounding box, objectness, and class predictions). The weights \(\lambda_{\text{bbox}}\), \(\lambda_{\text{obj}}\), and \(\lambda_{\text{cls}}\) correspond to the bounding box, objectness, and class components, respectively. 



\section{YOLOX}
\label{sec:yolox}

YOLOX is an anchor-free version of YOLO, with a simpler design, but better performance. It aims to bridge the gap between research and industrial communities \cite{DBLP:journals/corr/abs-2107-08430}.

In object detection models, the concept of \textit{anchors} is a principle that uses predefined bounding boxes all around the image before doing any detection.
It can be seen as a "pre-shot," which should improve the model's efficiency.
Because of its efficiency and accuracy for object detectors, most SOTA object detection models are anchor-based, such as Faster R-CNN, YOLO, and SSD.

However, this technique has several limitations, such as fixed anchor sizes and ratios, large number of anchors, sensitivity of anchor configuration, handling of overlapping anchors, localization accuracy, scale invariance, class imbalance, and training complexity.
Because of those limitations, recent object detection research focuses on anchor-free models such as YOLOX. 

\begin{figure}
    \centering
    \includegraphics[width=\linewidth]{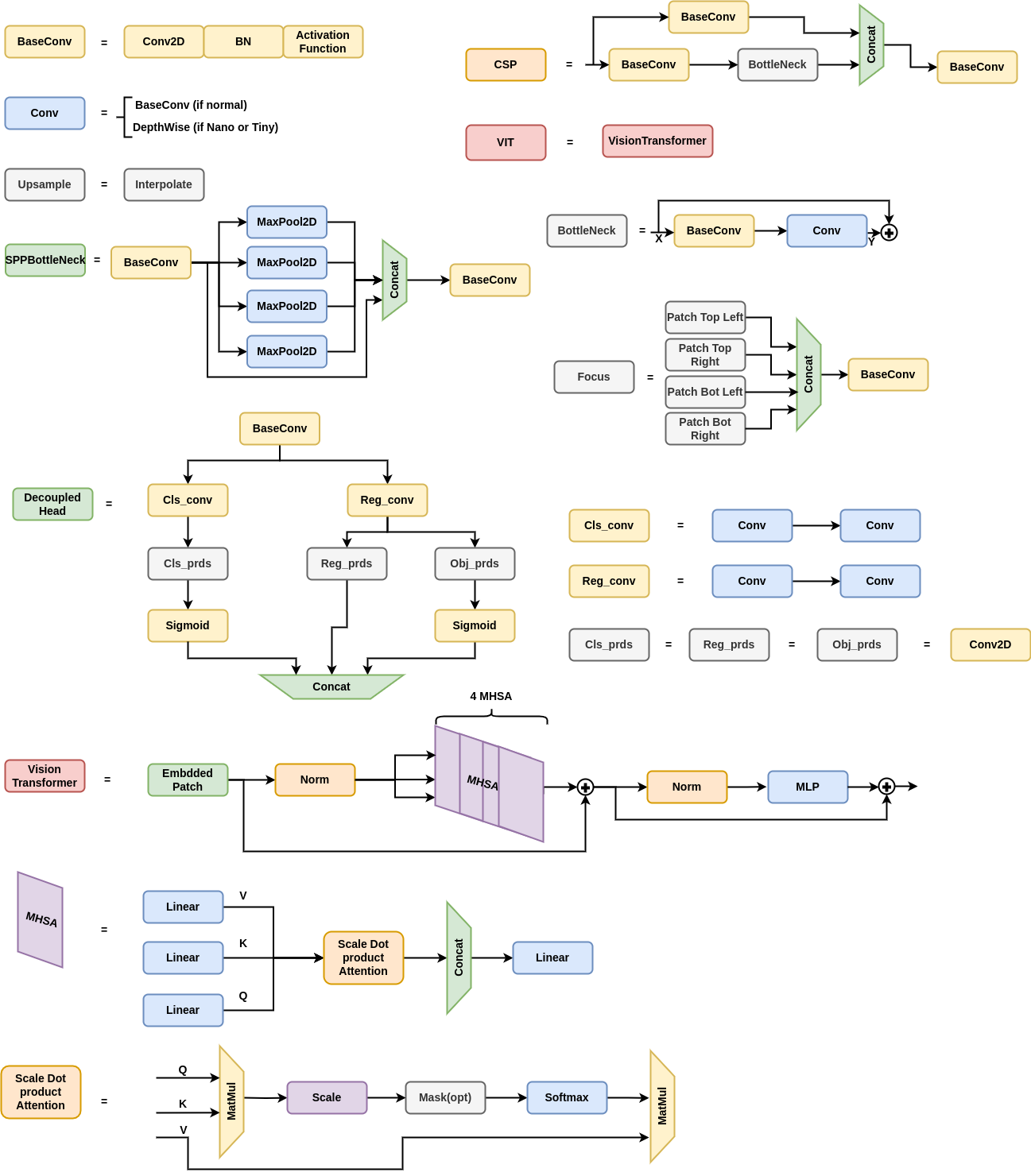}
    \caption{YOLOX-ViT Architecture additional description. Each module of Figure \ref{fig:yolox-vit} are explained on a lower level.}
    \label{fig:yolo-blocks}
\end{figure}

\section{Vision Transformer (ViT)}
\label{sec:transformers}

\noindent \textbf{Vision Transformer.}
In the Vision Transformer, an image is divided into patches, flattened, and linearly embedded into a sequence of tokens.
Let's assume an image \( I \) of size \( H \times W \times C \) is divided into \( N \) patches of size \( P \times P \times C \). Each patch is flattened and linearly projected to a \( D \)-dimensional embedding space. The process can be described as follows:
\begin{equation}
    \text{Patch}(I)_{(i, j)} = \text{Flatten}(I[iP:iP+P, jP:jP+P, :])
\end{equation}

\begin{equation}
    \text{Token}_{(i, j)} = \text{Patch}(I)_{(i, j)}W_E + b_E
\end{equation}

Where \( W_E \) is the embedding matrix of size \( P^2 \cdot C \times D \), and \( b_E \) is the bias vector.

\noindent \textbf{Scaled Dot-Product Attention.}
The Scaled Dot-Product Attention mechanism computes attention scores based on queries $Q$, keys $K$, and values $V$. The keys and queries are first multiplied and scaled by the square root of the dimension of the keys $d_k$. Then, the softmax function is applied to obtain weights multiplied by the values to produce the output. The equation is given by:
\begin{equation}
    \text{Attention}(Q, K, V) = \text{softmax}\left(\frac{QK^T}{\sqrt{d_k}}\right)V
\end{equation}

\noindent \textbf{Multi-Head Self-Attention (MHSA).}
MHSA runs multiple attention mechanisms (heads) in parallel. Each head applies the attention function to different queries, keys, and values projections. The outputs of these heads are then concatenated and multiplied by an output weight matrix $W^O$. The process is defined as:
\begin{equation}
    \text{MHSA}(Q, K, V) = \text{Concat}(\text{head}_1, \text{head}_2, \dots, \text{head}_h)W^O
\end{equation}
where each head is computed as:
\begin{equation}
    \text{head}_i = \text{Attention}(QW^Q_i, KW^K_i, VW^V_i)
\end{equation}
Here, $W^Q_i$, $W^K_i$, and $W^V_i$ are parameter matrices for each head $i$.

\noindent \textbf{Transformer Encoder.}
The Transformer Encoder includes a stack of identical layers consisting of two sub-layers: an MHSA and a position-wise feed-forward network. The output of each sub-layer is normalized with layer normalization. The layer is represented as:
\begin{equation}
    \text{LayerNorm}(x + \text{MHSA})
\end{equation}

\end{document}